# A Supervised Machine Learning Approach for Sequence Based Protein-protein Interaction (PPI) Prediction

Soumyadeep Debnath[1], Ayatullah Faruk Mollah[2(✉)]

[1] Tata Consultancy Services Limited, Kolkata, India
[2] Department of Computer Science and Engineering, Aliah University, Kolkata, India

`soumyadebnath13@gmail.com`[1], `afmollah@gmail.com`[2]

**Abstract.** Computational protein-protein interaction (PPI) prediction techniques can contribute greatly in reducing time, cost and false-positive interactions compared to experimental approaches. Sequence is one of the key and primary information of proteins that plays a crucial role in PPI prediction. Several machine learning approaches have been applied to exploit the characteristics of PPI datasets. However, these datasets greatly influence the performance of predicting models. So, care should be taken on both dataset curation as well as design of predictive models. Here, we have described our submitted solution with the results of the SeqPIP competition whose objective was to develop comprehensive PPI predictive models from sequence information with high-quality bias-free interaction datasets. A training set of 2000 positive and 2000 negative interactions with sequences was given to us. Our method was evaluated with three independent high-quality interaction test datasets and with other competitors' solutions.

## 1 Introduction

Protein works by interacting with other proteins to accomplish various biological processes essential to a living organism. Identification of such interactions facilitates different biological problems such as therapeutic target identification [10], characterization of metabolic and cellular activity, hormone regulation, signal transduction, DNA transcription and replication, drug target identification [7] etc. Protein-protein interactions (PPIs) detection has been improved with high throughput experimental technologies but on a small portion. These experimental approaches are expensive and time-consuming to apply over large-scale PPIs. Thus, there is immense need for reliable computational approaches to identify and characterize PPIs. The primary structure of a protein, represented by amino acid sequences is the simplest type of information that can be applied in PPI prediction through intuitive numerical feature representation. There are plenty of information in primary sequence of a protein and experimental studies reveal that the sequence is one of the key information for PPI prediction [14,11,13,6, 12]. The experimental studies become the evidence of applicability and efficacy of primary structure data in resolving different complex proteomic properties.

In recent studies [8,5,9], it has been observed that in most of the cases the PPI datasets used for evaluation were biased due to *component level overlapping*. For any PPI pair, sharing of individual component (sequence information) between training and test sets is referred to as *component-level overlapping*. For an unbiased and fair evaluation of the PPI predictive model, it is necessary to distinguish the test pairs on the basis of whether they share individual sequence with any pair of the training set [8]. The frequent occurrences of individual protein and/or pair in both training and test data raises the component level overlapping issue and bias the model greatly. To avoid this bias in PPI prediction, a dataset curation scheme is proposed by [8,5] where three levels of complex test classes are introduced for the predictive models. Park *et al.* [8] experimentally established that the performance of the top methods on unbiased datasets is significantly lower than their previously published results which raises the bar higher for PPI prediction methods. Here, the dataset was designed and curated by removing the component-level overlapping issue to make the PPI prediction unbiased and comprehensive.

## 2    Acknowledgement

*Sequence based Computational PPI prediction (SeqPIP-2020)*[1] at *International Conference on Frontiers in Computing and Systems (COMSYS-2020)*[2] was organised to build upon the high quality PPI dataset and encourage researchers across the world to introduce new techniques for sequence based generalized and unbiased PPI prediction. In sequence based PPI prediction, the main computational challenge is to find suitable way to describe important feature information. Moreover, dataset preparation is very crucial for machine learning methods. Here, unbiased training and test datasets are carefully curated and the competition rules are concocted in accordance with these objectives.

Through this competition, several sequence based approaches were introduced to improve the performances on the provided benchmark dataset. Our submission ranked **1st position** and a short intro with results comparison was already archieved in the competition draft paper [15]. In this paper, we described in details the dataset and our technique that performed well on the aforesaid challenging task. We hope that our algorithm submitted in the competition, described in this paper will be of use to the broader community of bioinformatics research.

---

[1] https://www.comsysconf.org/Comsys2020/seqpip2020.html
[2] https://www.comsysconf.org/Comsys2020/

## 3     Dataset Description

The positive interaction data is retrieved from HIPPIE ($v2:1$) [2] database and high quality subset of these datasets is selected for the competition. The high quality PPI data is selected based on confidence scoring (larger than 0.8) of HIPPIE. The negative protein-protein interaction (NPPI) data is generated by random sampling of protein pairs that are not known to interact in any benchmark PPI dataset. The sequence information is retrieved from UniprotKB/SwissProt database [3]. Protein sequences which are too similar should not occur simultaneously in training and test sets. Presence of redundant homologous sequences in PPI dataset greatly influences the performance of predictive models. To remove these homologous sequences, CD-Hit [1] is applied over the unique sequences that are present in the dataset with a threshold of 0.4 (40% identity). This ensures that, there is no issue of occurrences for identical sequences ($\leq 40\%$) that reside either between train/test or within train/test as pairs. Total 5435 unique non-redundant protein sequences are retrieved for final PPI dataset selection.

Finally, 4500 positive and 4500 negative interaction pairs are selected from previously extracted high confidence PPIs and randomly generated NPPIs respectively. All 9000 interactions are comprised by 5435 non-redundant unique sequences. To make the competition comprehensive and unbiased, the dataset is curated carefully and partitioned into training and test sets by removing component level overlapping as proposed by [8]. To implement the idea of removing *component-level overlapping* issue, test data is designed into three difficulty classes viz. C1, C2 and C3. The unbiased data preparation and detailed description of all three test classes (C1, C2 and C3) are discussed below.

A scheme is implemented similar to what was proposed by Park and Marcotte [8] to design the unbiased dataset for sequence based PPI prediction. For an unbiased and fair evaluation of the PPI predictive model, it is necessary to distinguish the test pairs on the basis of whether they share similar sequence information with the pairs of the train set. Here, the similarity is considered as over 40% identity between any two sequences. To overcome these issues, Park and Marcotte [8] proposed a scheme that, for any trainset the test cases will be partitioned into three distinct predictive test classes (C1, C2 and C3). In C1, both sequences of any test pair may be present in the trainset but not as a pair. In C2, only one component (sequence) can be present in the trainset and in C3, no components in the test pair could be present in the trainset. Detailed dataset curation and train/test partitioning is demonstrated with a toy example (with Graph based representation) in Fig. 1.

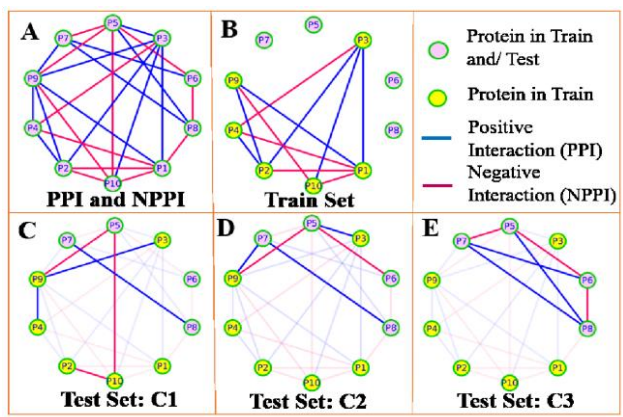

**Fig. 1.** Schematic diagram of data preparation. A) protein interaction network with PPI (red edges) and NPPI (blue edges). B) Selected Train set with the set of nodes (P1, P2, P3, P4, P9, P10), All the nodes in trainset are marked with yellow color (from B to E). C) Test set C1, with the set of nodes (P2, P3, P4, P9, P10), D) Test set C2, with the set of nodes (P3, P5, P6, P7, P8, P9). E) Test set C3, with the set of nodes (P5, P6, P7, P8). All three test classes ensure that they have not shared any exact pair (same edge) with the trainset. C1 has component level overlapping as well as pair (both components) sharing as both train and C1 shares P3, P2, P4, P9 and P10. For example, both proteins P3 and P9 from pair P3-P9 in C1 is also present in the trainset, and similarly for pair P2-P10, P4-P9. In C2, only one protein node is shared between train and test set. For example, in pair, P9-P5, P3-P5 only one node from each pair is shared such as P9, P3 respectively. In C3, no edges and nodes are shared between train and test set.

| Dataset Type | Total Interactions | Positive (PPI) | Negative (NPPI) | # of Unique Sequences |
|---|---|---|---|---|
| Train | 4000 | 2000 | 2000 | 3374 |
| C1 | 2000 | 1000 | 1000 | 1672 |
| C2 | 1500 | 750 | 750 | 978 |
| C3 | 1500 | 750 | 750 | 984 |
| Total | 9000 | 4500 | 4500 | 5435 |

**Table 1.** SeqPIP-2020 competition dataset details (C1, C2 and C3 are the three subsets of testset).

Total 4000 (2000 positive and 2000 negative) interactions were provided to the competitors as trainset. However, the test datasets were not shared to the competitors before announcement of results and the competitors' models were evaluated only once on the test datasets after the final submission of their models. Details of the competition datasets are shown in Table 1. The whole dataset (training and test sets) at three component levels viz. C1, C2 and C3 have been made publicly available[3] for academic, research and noncommercial purposes.

## 4    Methodological Process

Our primary objective was to predict (classify) the protein-protein interaction from a given dataset. As it has been already considered that if protein P1 can interact with protein P2 then the vise-versa is also possible. Therefore, from the training dataset we need to extract each of the *feature vectors* for every protein pair in such a way that;

1. It will be same for the protein pairs P1, P2 and P2, P1 both.
2. It should not impact the classifier to learn if the $2^{nd}$ occurrence of the protein pair is not available in the training dataset.

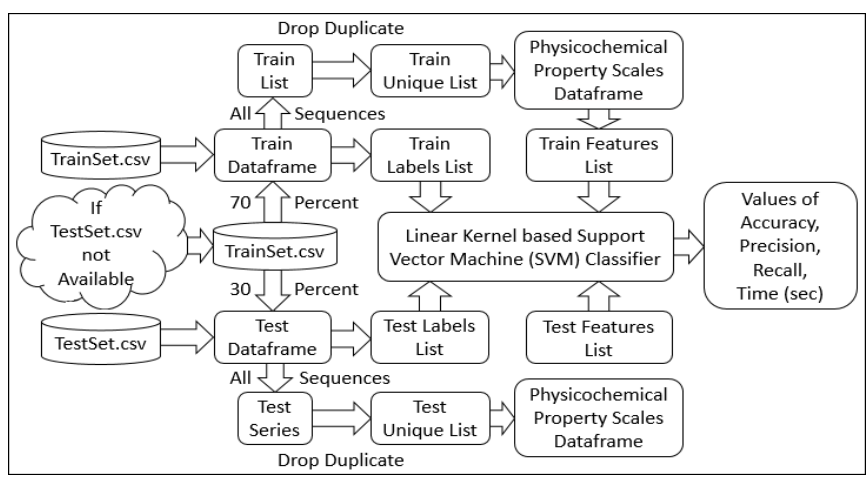

**Fig. 2.** Sequential Flow Diagram of Our Methodological Process

Here considering these scenarios, we have followed the above mentioned sequential methodological process (flow diagram in Fig. 2) in order to predict

---

[3] https://sites.google.com/site/bioinfoju/home/seqpip

*protein-protein interaction (PPI)* from the pairs of protein sequences of the provided dataset.

## 5  Feature Extraction: sequence-based Physicochemical Property Scales

Here for PPI prediction, firstly we selected all 20 essential amino acids (*A, C, D, E, F, G, H, I, K, L, M, N, P, Q, R, S, T, V, W, Y*) based on which the protein sequences are formed and characterized the proteins by 12 physicochemical properties of their composite amino acids [16–23] namely, *hydrophilicity (H2), flexibility (F), accessibility (A1), turns scale (T), exposed surface (E), polarity (Pa), antigenic propensity (A2), hydrophobicity (Ha), net charge index of the side chains (NCI), polarizability (P2), solvent accessible surface area (SASA), and side-chain volume (V)*. Among the 12 properties, *hydrophobicity* and *polarity* were calculated according to two different scales or methods like *H11a, H12a* and *P11a, P12a* respectively. The values of 14 physicochemical property scales of the 20 essential amino acids have been listed below in the table of Fig. 3.

| AA | $H_{11}$[a] | $H_{12}$[a] | $H_2$ | NCI | $P_{11}$[a] | $P_{12}$[a] | $P_2$ | SASA | V | F | $A_1$ | E | T | $A_2$ |
|---|---|---|---|---|---|---|---|---|---|---|---|---|---|---|
| A | 0.62 | 2.1 | −0.5 | 0.007 | 8.1 | 0 | 0.046 | 1.181 | 27.5 | −1.27 | 0.49 | 15 | −0.8 | 1.064 |
| C | 0.29 | 1.4 | −1.0 | −0.037 | 5.5 | 1.48 | 0.128 | 1.461 | 44.6 | −1.09 | 0.26 | 5 | 0.83 | 1.412 |
| D | −0.9 | 10.0 | 3.0 | −0.024 | 13.0 | 40.7 | 0.105 | 1.587 | 40.0 | 1.42 | 0.78 | 50 | 1.65 | 0.866 |
| E | −0.74 | 7.8 | 3.0 | 0.007 | 12.3 | 49.91 | 0.151 | 1.862 | 62.0 | 1.6 | 0.84 | 55 | −0.92 | 0.851 |
| F | 1.19 | −9.2 | −2.5 | 0.038 | 5.2 | 0.35 | 0.29 | 2.228 | 115.5 | −2.14 | 0.42 | 10 | 0.18 | 1.091 |
| G | 0.48 | 5.7 | 0.0 | 0.179 | 9.0 | 0 | 0 | 0.881 | 0 | 1.86 | 0.48 | 10 | −0.55 | 0.874 |
| H | −0.4 | 2.1 | −0.5 | −0.011 | 10.4 | 3.53 | 0.23 | 2.025 | 79.0 | −0.82 | 0.84 | 56 | 0.11 | 1.105 |
| I | 1.38 | −8.0 | −1.8 | 0.022 | 5.2 | 0.15 | 0.186 | 1.81 | 93.5 | −2.89 | 0.34 | 13 | −1.53 | 1.152 |
| K | −1.5 | 5.7 | 3.0 | 0.018 | 11.3 | 49.5 | 0.219 | 2.258 | 100 | 2.88 | 0.97 | 85 | −1.06 | 0.93 |
| L | 1.06 | −9.2 | −1.8 | 0.052 | 4.9 | 0.45 | 0.186 | 1.931 | 93.5 | −2.29 | 0.4 | 16 | −1.01 | 1.25 |
| M | 0.64 | −4.2 | −1.3 | 0.003 | 5.7 | 1.43 | 0.221 | 2.034 | 94.1 | −1.84 | 0.48 | 20 | −1.48 | 0.826 |
| N | −0.78 | 7.0 | 2.0 | 0.005 | 11.6 | 3.38 | 0.134 | 1.655 | 58.7 | 1.77 | 0.81 | 49 | 3.0 | 0.776 |
| P | 0.12 | 2.1 | 0.0 | 0.240 | 8.0 | 0 | 0.131 | 1.468 | 41.9 | 0.52 | 0.49 | 15 | −0.8 | 1.064 |
| Q | −0.85 | 6.0 | 0.2 | 0.049 | 10.5 | 3.53 | 0.18 | 1.932 | 80.7 | 1.18 | 0.84 | 56 | 0.11 | 1.015 |
| R | −2.53 | 4.2 | 3.0 | 0.044 | 10.5 | 52.0 | 0.291 | 2.56 | 105 | 2.79 | 0.95 | 67 | −1.15 | 0.873 |
| S | −0.18 | 6.5 | 0.3 | 0.005 | 9.2 | 1.67 | 0.062 | 1.298 | 29.3 | 3.0 | 0.65 | 32 | 1.34 | 1.012 |
| T | −0.05 | 5.2 | −0.4 | 0.003 | 8.6 | 1.66 | 0.108 | 1.525 | 51.3 | 1.18 | 0.7 | 32 | 0.27 | 0.909 |
| V | 1.08 | −3.7 | −1.5 | 0.057 | 5.9 | 0.13 | 0.14 | 1.645 | 71.5 | −1.75 | 0.36 | 14 | −0.83 | 1.383 |
| W | 0.81 | −10 | −3.4 | 0.038 | 5.4 | 2.1 | 0.409 | 2.663 | 145.5 | −3.78 | 0.51 | 17 | −0.97 | 0.893 |
| Y | 0.26 | −1.9 | −2.3 | 117.3 | 6.2 | 1.61 | 0.298 | 2.368 | 0.024 | −3.3 | 0.76 | 41 | −0.29 | 1.161 |

$H_{11}$ & $H_{12}$ hydrophobicity, $H_2$ hydrophilicity, NCI net charge index of side chains, $P_{11}$ & $P_{12}$ polarity, $P_2$ polarizability, SASA solvent-accessible surface area, V volume of side chains, F Flexibility, $A_1$ Accessibility, E Exposed, T Turns, $A_2$ Antegenic
[a]Hydrophobicity ($H_{11}$ & $H_{12}$) and polarity ($P_{11}$ & $P_{12}$) were calculated by two different methods

**Fig. 3.** Values of 14 physicochemical property scales of 20 amino acids

We translated each amino acid into a vector of 14 numeric values, each corresponding to a physicochemical scale value in the table of Fig. 3. As proteins have different lengths with different numbers of amino acids, therefore, we calculated the mean value for each physicochemical property scale to generate symmetric feature vectors for all protein sequences with varying numbers of amino acids.

For example, Fig. 4 shows the transformation of two protein sequences in a pair, P1 (AYCRS) and P2 (HRS), into 14-element feature vectors. Each element in each vector corresponds to a physicochemical scale value [24, 25].

| | Scale | $AA_1 = A$ | $AA_2 = Y$ | $AA_3 = C$ | $AA_4 = R$ | $AA_5 = S$ | Mean | | Scale | $AA_1 = H$ | $AA_2 = R$ | $AA_3 = S$ | Mean |
|---|---|---|---|---|---|---|---|---|---|---|---|---|---|
| | $H_{11}$ | 0.62 | 0.26 | 0.29 | -2.53 | -0.18 | -0.31 | | $H_{11}$ | -0.4 | -2.53 | -0.18 | -1.04 |
| | $H_{12}$ | 2.1 | -1.9 | 1.4 | 4.2 | 6.5 | 2.46 | | $H_{12}$ | 2.1 | 4.2 | 6.5 | 4.27 |
| | $H_2$ | -0.5 | -2.3 | -1 | 3 | 0.3 | -0.1 | | $H_2$ | -0.5 | 3 | 0.3 | 0.94 |
| | NCI | 0.007 | 117.3 | -0.037 | 0.044 | 0.005 | 23.46 | | NCI | -0.011 | 0.044 | 0.005 | 0.013 |
| | $P_{11}$ | 8.1 | 6.2 | 5.5 | 10.5 | 9.2 | 7.9 | | $P_{11}$ | 10.4 | 10.5 | 9.2 | 10.03 |
| $P_1$: | $P_{12}$ | 0 | 1.61 | 1.48 | 52 | 1.67 | 11.35 | $P_2$: | $P_{12}$ | 3.53 | 52 | 1.67 | 19.07 |
| | $P_2$ | 0.046 | 0.298 | 0.128 | 0.291 | 0.062 | 0.165 | | $P_2$ | 0.23 | 0.291 | 0.062 | 0.19 |
| | SASA | 1.181 | 2.368 | 1.461 | 2.56 | 1.298 | 1.77 | | SASA | 2.025 | 2.56 | 1.298 | 1.961 |
| | V | 27.5 | 0.024 | 44.6 | 105 | 29.3 | 41.28 | | V | 79 | 105 | 29.3 | 71.1 |
| | F | -1.27 | -3.3 | -1.09 | 2.79 | 3 | 0.026 | | F | -0.82 | 2.79 | 3 | 1.66 |
| | $A_1$ | 0.49 | 0.76 | 0.26 | 0.95 | 0.65 | 0.622 | | $A_1$ | 0.84 | 0.95 | 0.65 | 0.81 |
| | E | 15 | 41 | 5 | 67 | 32 | 32 | | E | 56 | 67 | 32 | 51.67 |
| | T | -0.8 | -0.29 | 0.83 | -1.15 | 1.34 | -0.014 | | T | 0.11 | -1.15 | 1.34 | 0.1 |
| | $A_2$ | -1.064 | -1.161 | -1.412 | -0.873 | -1.012 | 1.104 | | $A_2$ | -1.105 | -0.873 | -1.012 | 0.997 |

**Fig. 4.** Feature Extraction for each Protein Sequence of a Protein Pair test set.

After extraction of each protein features from a protein pair, in order to generate a standard and simple feature vector for each protein pair we calculated the mean value among the two feature vectors. This process was repeated for all the protein pairs present in the dataset.

## 6 Model/Classifier

After successful completion of the feature extraction stage for all the protein pairs from both the train and test samples of the dataset, we applied all feature lists and label lists of both train and test data to *linear kernel* based *support vector machine (SVM)* classifier for the classification or prediction of protein protein interaction (PPI).

## 7 Results and Discussion

We calculated the performance of our this method of SeqPIP tested on the three different test datasets and the results are reported in Table 2. The results were also validated by *k-Fold Cross-Validation* with k=10. Different statistical metrics such as precision, recall, accuracy and F1-score were computed to evaluate the models. Our method standed with highest performance in terms of statistical metrics and also time complexity.

| Test Class | Precision | Recall | Accuracy | F1-score | Rank |
|---|---|---|---|---|---|
| C1 | 0.667 | 0.623 | 0.634 | 0.644 | |
| C2 | 0.632 | 0.603 | 0.611 | 0.617 | 1 |
| C3 | 0.524 | 0.581 | 0.551 | 0.527 | |

**Table 2.** Performance evaluation of the best method of SeqPIP.

Detailed comparisons of other standard methods were mentioned in the competition archieved version paper [15]. The nonML based approaches are not comprehensive in the performance evaluation metric compared to ML based methods and far away from.

# 8 Conclusion & Future Scope

Here, we have presented our sequence based PPI prediction approach and it's performance in connection with our result at *SeqPIP-2020* competition. Different machine learning based and non-machine learning based approaches were employed to predict the PPIs by exploring the primary sequence information of proteins. All the methods were trained and evaluated on unbiased train and test data sets by removing *component level overlapping* biases. Among the best four approaches [15], the overall performance of machine learning based approaches were found superior than other approaches and our supervised approach was recognized as the best machine learning approach among all submissions. Our prediction model can be more improved with more efficient feature extraction techniques on which we are currently working on, therefore, we have also worked on biomolecular clustering techniques for PPI networks [4]. The datasets and the methods of SeqPIP enabled a platform for many bioinformatics applications such as complex detection, characterisation of functional relationship, PPI network analysis, etc.